# A Novel Robot-Assisted Learning-by-Teaching Pedagogy for Children with ASD

Laura Boccanfuso[1]*, Erin Barney[1], Marilena Mademtzi[1], Claire Foster[1], Quan Wang[1], Colette Torres[1], Lisa Chen[1], Brian Scassellati[2], Pamela Ventola[1], Frederick Shic[1]§

[1]Yale Child Study Center, Yale School of Medicine, New Haven, CT, USA
[2]Yale Social Robotics Lab, Department of Computer Science, Yale University, New Haven, CT, USA
*Now at Van Robotics, Inc.
§Now at Seattle Children's Hospital

## ABSTRACT

Interaction paradigms used in robot-assisted autism intervention have historically employed robots as teachers, clinical assistants, or more-abled peers to promote a variety of social skills. These modalities often leverage the expertise of trained practitioners to ensure that child-robot interactions are productive or clinically grounded to yield positive therapeutic benefits for children across the autism spectrum. Yet, despite the fact that the majority of children with autism spectrum disorder (ASD) attend mainstream schools and spend 80% or more of their time in the general classroom [27], there is a paucity of research incorporating validated classroom teaching pedagogies into robot-assisted autism interventions.

*In this work, we introduce a novel teaching methodology for advancing social skills in school-aged children with ASD.* We evaluate the effectiveness of a novel robot-assisted autism intervention which incorporates the learning-by-teaching pedagogy and explores the comparative benefits of employing a robot versus a human confederate for improved performance on a set of social skills tasks. Results show that 80% of study participants performed better in the robot condition (mean performance in the robot condition=63%, mean performance in the confederate condition=37%), irrespective of the scenario order. Further, 90% of all participants were significantly more engaged in the robot condition (mean engagement: robot=61%, confederate=32%) and, while the effect did not result in the confederate condition, analyses indicate that overall engagement in the robot condition contributed to improved performance.

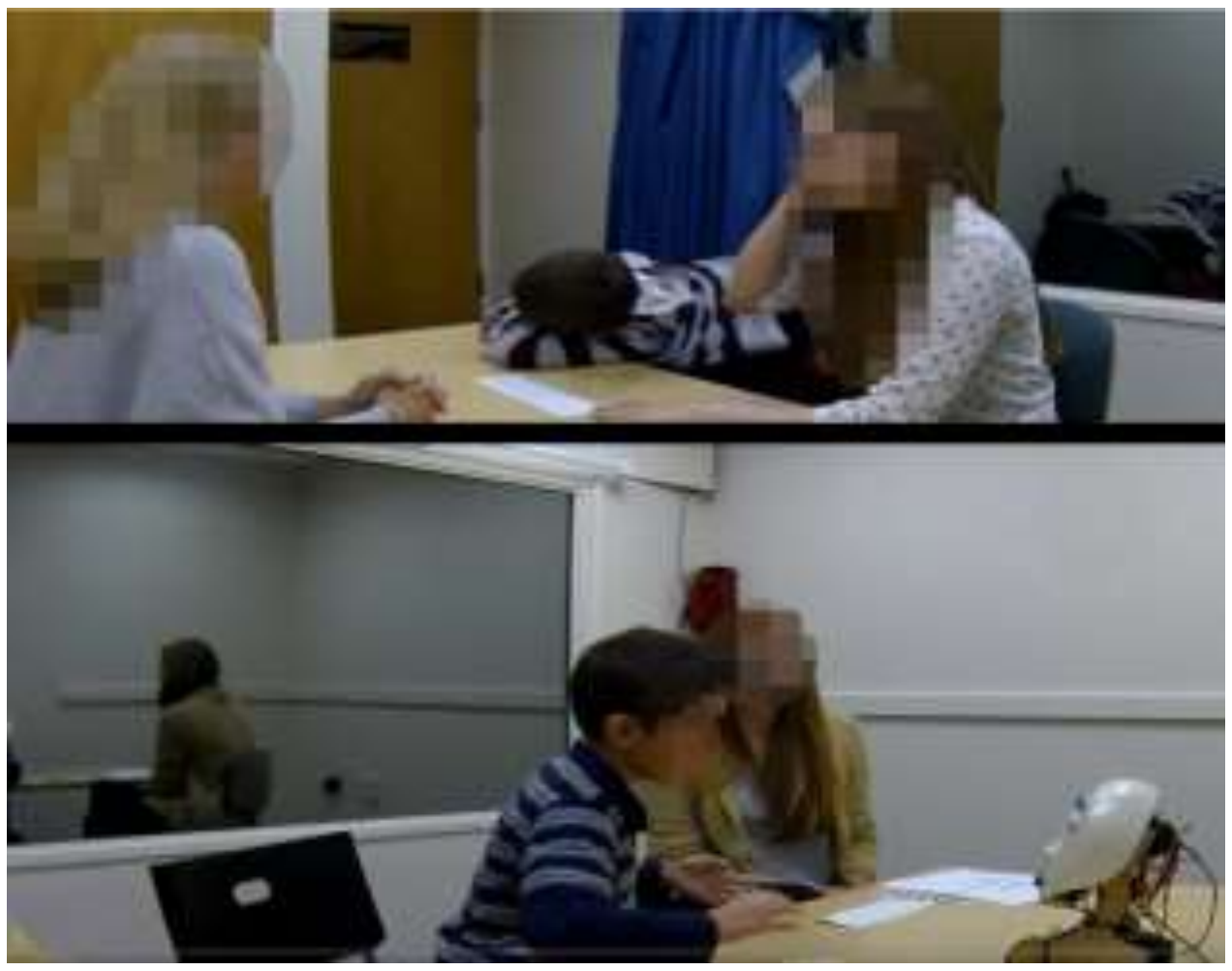

**Figure 1:** Top: Participant in confederate condition. Bottom: Participant in robot condition.

These results suggest that robots employed in a learning-by-teaching context may help enhance engagement and improve performance on a simple social skills task for children with ASD.

# 1 INTRODUCTION

Interventions for promoting social skills in children with autism often include social scenario modeling followed by discussion which highlights particular social missteps that may be targeted in the intervention. Some of the most recently validated techniques include video modeling, peer-facilitated academic tutoring and group-based social skills interventions [34]. Directly enlisting peers in interventions has been found to be very effective and there is ample literature describing immediate and long-term benefits of peer-enabled academic tutoring including improved understanding of material and increased confidence [22].

One instantiation of peer-facilitated intervention is known as the learning-by-teaching pedagogy, or peer tutoring, and is established as a one-on-one teaching process in which “the teacher” is a peer possessing a similar academic status as “the student”. In this methodology, the peer teacher may focus on conveying techniques for advancing academic performance or on engaging his/her peer in interpersonal exchanges to promote specific social skills [24]. These approaches are supported by mounting evidence that social skills play a critical role in school adjustment and subsequent academic performance [28, 29, 30].

A recent review of peer tutoring studies involving children with autism, reports that while preparing to teach helped students better manage essential processing, the actual act of teaching fostered deeper generative processing, which is critical for achieving long-term learning [19, 24]. This is particularly true for low-achieving students [23]. Yet, intervention paradigms focusing on social skills peer tutoring by children with ASD are scarce.

Several obstacles complicate the deployment of this type of paradigm for children with autism and their peers. Some of the most significant challenges include: (i) limited opportunities to find and match age-appropriate peers with similar social skills challenges within the same class, (ii) the lack of necessary motivation for the child to take on the role of teacher, particularly if it involves additional effort during social interactions with another child and, (iii) inexperience: where the child with autism (in the “teacher role”) has not yet mastered the social skill and may provide inappropriate feedback.

Social robots offer numerous functional benefits for social skills peer tutoring in children with autism. Robot interactions can be customized to present a specific set of social skills which are particularly challenging for each child. Further, the added social complexity inherent in one-on-one interactions with other peers is minimized, allowing the child to focus on the specific social skill of interest. Finally, the peer teacher’s instruction and feedback from the robot “student” upon receiving a correction occurs in a controlled setting. This scenario allows the peer teacher time to practice his/her approach without the potential repercussions of an actual social exchange gone awry.

For these reasons, employing a social robot in the role of the peer “student” may be uniquely advantageous for exploring the potential efficacy of a robot-assisted learning-by-teaching approach to augment existing approaches that promote social skills in children with autism. In this first study, we evaluate the efficacy of a robot-assisted, learning-by-teaching approach through a simple, social skills task with 10 children with autism in a randomized control trial. We invited 10 children with ASD, with varying cognitive ability, to participate in this study and examined their task performance in correctly identifying four social skills errors. The primary social skills task includes identifying and correcting four fundamental social missteps while observing a conversational dialogue between the session facilitator and the human confederate or robot

# 2 BACKGROUND

## 2.1 A Brief Overview of the Generative Process of Learning

The authors in [19] present evidence in their review demonstrating that the actual act of teaching fosters deeper generative processing and long term learning. The underlying premise of these findings is better appreciated with a general understanding of the generative process. A primary tenet of the generative model of learning is that individuals tend to generate perceptions and meanings that are congruent with their own prior learning. What this key principle signifies for long-term learning, is that it inherently involves generating associations between learning stimuli and stored information. Substantial evidence also shows, at least in the scope of education, that even young children have firmly developed views which can be in conflict with what educators believe, and these perceptions may not change even with instruction [32, 33]. Therefore, the generative model for learning actively recognizes the importance of integrating the learner’s perception and interpretation of stimuli while actively constructing meaning from it [34].

Learning-by-teaching methodologies provide opportunities for peer tutors to integrate their own experiences into the teaching process to further deepen their own understanding in ways that build confidence and reinforce concepts which facilitate long-term learning. School-aged children with autism have often received considerable intervention and therapy through their school or from after-school therapies, but their perspectives on how to impart what they’ve learned can differ widely. For these reasons, it is critical to encourage children to incorporate their own frame of reference into the learning process. In so doing, we may also advance our own understanding of how individual children with autism process the appropriate use of social skills. Peer tutoring inherently creates opportunities to learn those perspectives.

## 2.2 Leveraging key robot features to optimize generative learning

Socially assistive robotics (SAR) research towards the promotion of therapeutic and educational goals has made great progress in the past decade. A critical review of research in the field concludes that children with autism exhibit better understanding of objects compared to the social world, are more responsive to feedback delivered by technology and are more intrinsically interested in electronic or robotics components [1].

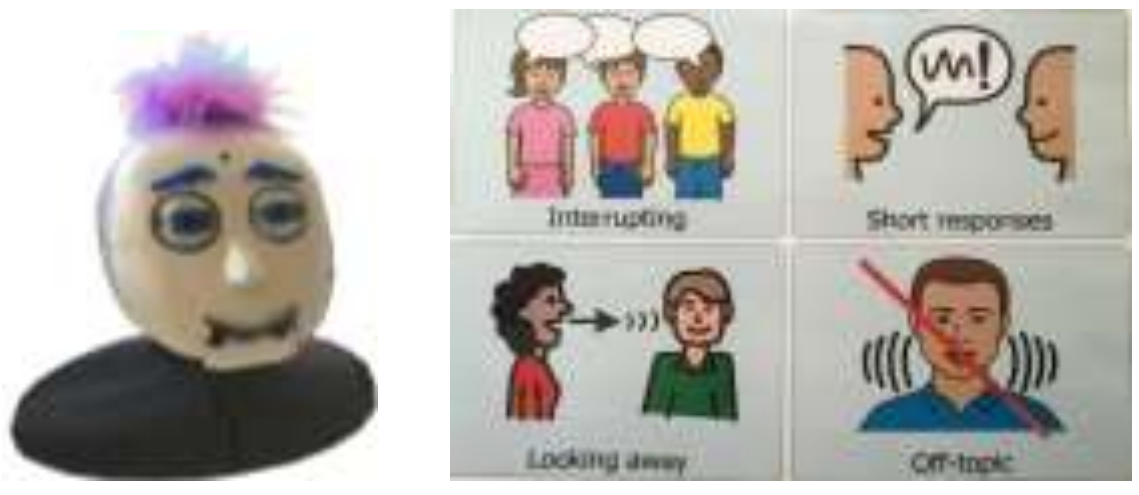

**Figure 2**: Left: The robot named L-E. Right: The social skills chart provided to each participant.

SAR studies have reported social and communicative skills improvements in children with autism, although the variety of robot-assisted methodologies employed are as diverse as the types of robots used and the skills targeted [6, 8]. Research focusing on younger and lower functioning children with autism typically addresses the promotion of early-emerging, core competencies such as shared attention, spontaneous imitation and turn-taking [13, 29, 30] whereas others examine the impact of child-robot interactions on the incidence of other important social deficits associated with autism including speech, eye gaze, affective expression and related social behaviors contributing to engagement [6, 10, 27, 28, 31].

Early studies employing robots to assist children with autism have revealed the breadth of potential benefits robot-assisted paradigms may offer this population [18, 20, 22]. These foundational studies point to key insights regarding social robot attributes that can be leveraged to benefit children with social and communicative deficits. Among these are the potential for the promotion of theory of mind [35] and prosocial behaviors [10] through targeted interventions with robots. In the past decade, research in the field has accelerated, with increasing effort being dedicated to more fully exploring the hardware, interaction modalities and autonomy of systems deployed for this purpose [23, 24, 25].

Engagement is a particularly important aspect of intervention for children with ASD [6], and considerable research exists supporting the positive effect of social robots on engagement within this population [36, 37]. It is widely believed that robots are comparatively less complex and less socially demanding than humans, thereby reducing the social burden on children with ASD and contributing to increased engagement [13,36].

Social robots are also highly customizable and can accommodate a wide range of skill proficiencies and personalities. They can respond in ways that are designed to be constructive and reinforcing, regardless of previous behavior that might otherwise negatively impact a social exchange. Robots can also be explicitly designed to focus on specific communication or social targets and repeat enacted social scenarios without fatigue or frustration. Further, social robots can fulfill a variety of social roles, ranging from authoritative to that of a lesser- or similarly-abled peer.

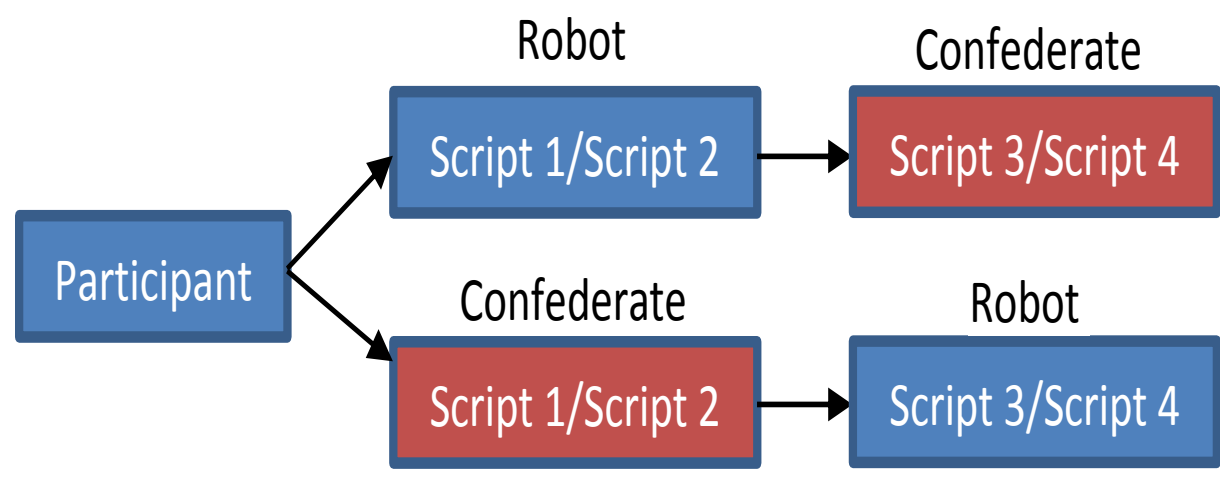

**Figure 3**: Randomized study condition order.

## 3 METHODOLOGY

We invited 10 children diagnosed with autism to participate in two interactive sessions designed to examine performance on a simple, social skills task. This randomized control study was designed to quantitatively compare performance differences of a group of school-aged children with ASD while using a robot and a human confederate to identify and correct four social missteps. Our approach was inspired by the learning-by-teaching pedagogy in which common social mistakes were described to each child and the child was provided numerous opportunities to identify and correct them as they occurred during structured interactions.

One session (the confederate condition) included the child, a human facilitator, and a human confederate and the other session (the robot condition) included the child, a human facilitator and a robot. The duration of the study protocol for both conditions was approximately 30 minutes, and included the same seven phases. The length of each phase was timed to ensure consistency across both conditions. The only notable distinction, in addition to the presence of either a robot or human confederate, was the actual script for the stories and the content of scripts in each session. In order to mitigate ordering effects that may have contributed to differences in performance between the robot and confederate conditions, the order of conditions presented to each participant was randomized.

### 3.1 Robot

A custom-designed robot prototype was developed by modifying a commercially available robot kit [16] to provide a simplified interface with sufficiently articulated facial features to provide expressivity and sustained attentional focus (Fig. 2).

The robot features an articulated neck, mouth, eyes and eyebrows, and a torso that can be secured to a desk to provide stability and enhance safety. The robot's articulated mouth has two degrees of freedom (DOF) to facilitate smiling, frowning and phoneme-synchronized speech. The robot also features three DOF in the eyes: one DOF in the eyelids for blinking, and two DOF in the eyebrows to support the expression of happiness and sadness. Finally, there are two DOF in the neck to allow for the robot to direct its interactions to the facilitator and the child.

To ensure robust, natural interactions, all free conversation with the robot was teleoperated and supported by the commercial text-to-speech software, IVONA. Stories and scripts included in the study protocol were programmed a priori to ensure consistent timing and delivery across participants.

## 3.2 Participants

Ten children between 6-10 years of age (*m=8.75 yrs.*) who were diagnosed with autism based on evaluation with gold standard DSM V criteria and met or surpassed a minimum threshold (>=80) for IQ, as determined by the Differential Ability Scales-General Conceptual Ability (DAS-GCA) assessment (*m=107.5*), were invited to participate.

## 3.3 Study design

Each child was randomly assigned to one of two study groups, a confederate-first or a robot-first group (Fig. 3). The study protocol was briefly described to the child and the facilitator led the child to the experiment room with 3 chairs, a table, and either a confederate or a robot. (Fig. 1). Either the confederate or robot greeted the participant and introduced her/itself. The remainder of the session proceeded as illustrated in Table 1.

During Phase 1, the robot/confederate, child and facilitator had the opportunity to engage in unstructured conversation. This phase was included to allow the robot/confederate to introduce itself/herself and allow the child time to acclimate to the interaction scenario. In Phase 2, the robot/confederate tells a short fictional story to the child while modelling appropriate social behavior (making eye contact, staying on topic) and encourages the child to tell their own story in Phase 3. Phase 4 features a second short story, again enabling the robot/confederate the opportunity to model appropriate social behavior.

At the conclusion of Phase 4, the facilitator provided the participant with a laminated page illustrating each of the four social missteps (Fig. 2) and then encouraged the child to select one large buzzer-button. Next, during each participant's first session only, the facilitator described how to play the script game and guided the child through a demonstration. The child's role was to observe a pre-scripted dialogue between the facilitator and the robot/confederate. Upon observing the robot/confederate make a social error, the child would hit the buzzer and the conversation was paused. The facilitator would ask the child which misstep s/he observed and what the robot/confederate should do next time to fix it. The game continued until each scripted dialogue was completed. The facilitator provided as much guidance as necessary to identify the correct answer when the corresponding error was presented but only during the demonstration script. During the remaining scripts, no guidance, positive or negative feedback about the correctness of the child's answer was given.

**Table 1:** Study Protocol

| Phase | Description |
|---|---|
| 1 | Free conversation with facilitator, confederate/robot and child. 5 minutes. |
| 2 | Confederate/robot tells story 1 |
| 3 | Confederate/robot invites child to tell a story |
| 4 | Confederate./robot tells story 2 |
| 5 | (Game demo) Script 1 game |
| 6 | Script 2 game |
| 7 | Free conversation with facilitator, confederate/robot and child. 3 minutes. |

**Four social skills selected**. In this study, we developed a simple social skills task to focus on four social skills errors commonly made by this population. These included: (i) Looking away while speaking, (ii) Interrupting, (iii) Abrupt changing the topics topic of conversation and, (iv) Short responses. Each of these social skills errors were selected because they are common social skills errors made by this population, each is discretely quantifiable and easily simulated with the robot developed for this study.

# 4 DATA

## 4.1 Data collection

Video and audio data were collected for each 30-minute session in order to record performance on each task, estimated head and gaze direction, physical movements, and vocal utterances for each participant and in both study conditions.

## 4.2 Data analysis

Each session video was manually coded to score the frequency, quality of participant responses and engagement with social targets. A two-way mixed, single measures, inter-rater correlation coefficient (ICC) was computed and a minimum ICC of 75% across four coders for 20% of videos coded was achieved across two features including the: (1) number of correctly identified social missteps: measured as a ratio of correctly identified social missteps/opportunities offered for each script and, (2) quality of social misstep solutions provided. A point scale from 0-2 was awarded for the quality of the solution provided for each correctly identified misstep. Answers that were irrelevant to the error received 0 points, scripted answers (e.g., repeating the same wording of the social error but adding "not") received 1 point. Relevant, specific responses received 2 points.

In addition, the duration of directed attention to each of three social targets was manually coded by a single coder, not previously familiar with the study or its outcome measures. Durations for attention to the confederate, the facilitator and robot were annotated as a percentage consisting of the total looking time at the target divided by the total time for a specific phase portion. This calculation effectively ensures that small variances in phase times did not unduly affect attention durations for each participant. Additionally, by coding attention to each target as a percentage of the total time coded, we were also able to evaluate the total time participants spent looking at any social target.

**Performance**. Performance measures were calculated as the percentage of correct answers provided for Script 1, percentage of correct answers for Script 2 and the percentage of total correct answers, within each study condition. Due to very few unanticipated interruptions or system delays, there were a few instances when the actual number of robot- or confederate-elicited social missteps differed slightly from the number of planned opportunities. For this reason, a ratio of number of correctly identified social missteps to the total number of missteps presented was used to compute performance.

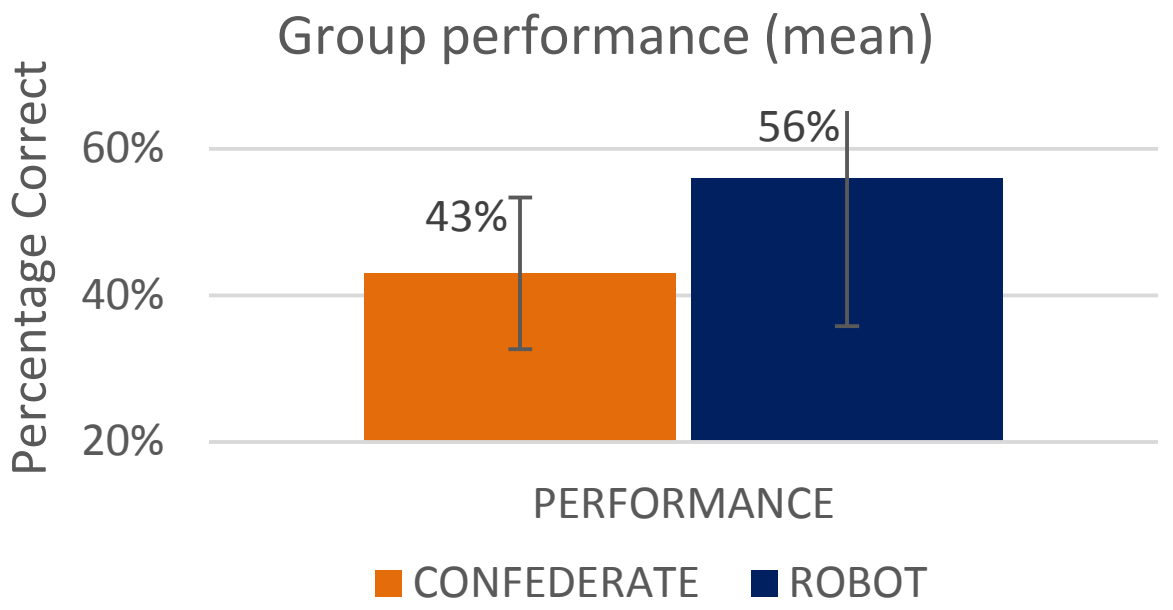

**Figure 4**: Higher overall group performance in the robot condition.

**Engagement**. Four social engagement targets were identified and evaluated for this study, including: (1) Confederate, (2) Robot, (3) Facilitator in the confederate condition, and (4) Facilitator in the robot condition. All measures of engagement were computed as a percentage of absolute looking time, defined as the time spent attending to the target divided by the total time coded for that session part. Since interactions were more were structured (and inherently less natural) during Phases 2-6, one 5-minute window at the beginning of Phase 1 and one 3-minute window at the beginning of Phase 7 were manually coded to calculate engagement durations. Due to the less structured format of free conversation, some participants engaged in slightly less or more time than what was allotted for Phases 1 and 7, so the exact times are approximate. Because engagement measures are a ratio of the time engaged with a specific target divided by the total time engagement was measured, slight variations in the total coded time windows did not directly affect results and were allowed. In this way, we were able to not only evaluate the absolute total during each condition, but also the total time spent looking at *any* social target. Our analyses include an individual engagement measure for each target within each condition, and an overall engagement measure in each condition.

We computed Pearson bi-variate correlations for each of the performance and engagement measures to evaluate potential connections between performance scores and attention durations within and between study conditions. We also performed Fisher's z-transformation on within- and between-condition correlations to investigate the significance of differences between correlations. Finally, we conducted multiple linear regression to examine the predictive value of engagement, age and IQ for predicting performance in both the robot and confederate conditions.

# 5 RESULTS

The two primary objectives of this study were to: (1) evaluate the effectiveness of a robot employed in a novel learning-by-teaching interaction paradigm for a group of children with ASD and, (2) examine the comparative benefits of using a robot compared to a human confederate in this context. In so doing, we present analyses measuring performance and engagement

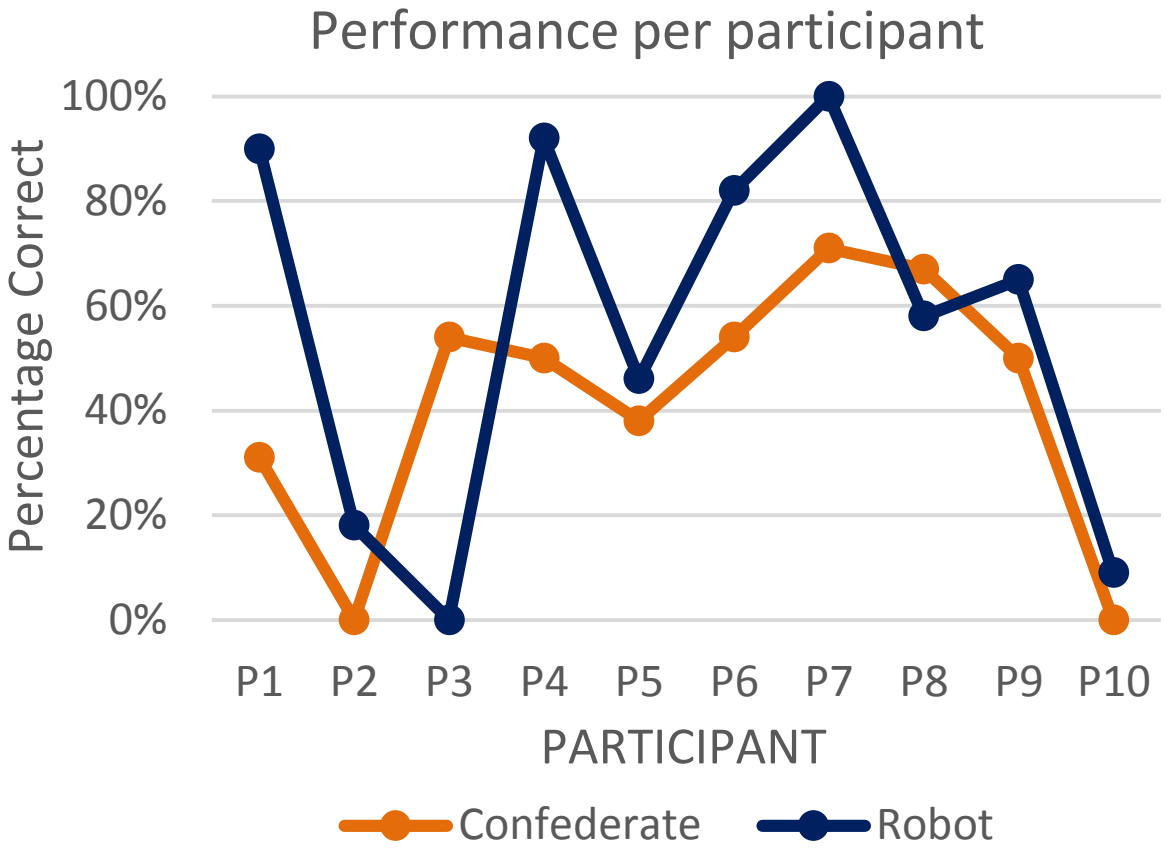

**Figure 5**: Per participant performance showing 80% of participants achieved better performance in robot condition.

differences by condition and IQ level for both robot and human confederate study conditions.

## 5.1 Performance results

### 5.1.1 *Performance by condition.*

**Between conditions.** Overall mean task performance in the robot condition exceeded overall mean performance in the confederate condition (Fig. 4). The performance mean in the robot condition was 56%, ranging from 0% to 100%. Comparatively, the mean performance in the confederate condition was 43%, ranging from 0% to 71%. Further, a full 80% of participants performed better in the robot condition, with performance improvements ranging from 8% to 60% and averaging approximately 26% overall (Fig. 5). No significant correlation resulted between performance in the confederate condition and the robot condition, nor between performance in either Script 1 or Script 2 for both conditions.

**Within conditions.** By contrast, a very strong correlation between performance in Script 1 and Script 2 in the confederate condition resulted ($r=0.965$, $p<0.01$). Similarly, a strong correlation was found between performance in Script 1 and Script 2 in the robot condition ($r=0.787$, $p<0.01$). These correlations indicate that the difficulty level for scripts were similar and their order would not have likely affected performance. An examination of scores from individual scripts within the robot condition reveals that 80% of participants either maintained or improved performance from Script 1 to Script 2, with an average increase of approximately 11%. By comparison, 60% of participant scores from scripts in the confederate condition either stayed the same or improved from Script 1 to Script 2, representing an average gain of approximately 8%.

### 5.1.2 *Performance by IQ.*

**Between conditions**. Verbal IQ and IQ were significantly correlated with overall performance in the confederate condition ($r=0.794$, $p<0.05$ and $r=0.726$, $p<0.05$, respectively), and Script 2 ($r=0.741$, $p<0.05$) and strongly correlated in Script 1 ($r=0.843$, $p<0.01$). Interestingly, no correlation resulted in the robot condition between Verbal IQ/IQ and overall task performance or for any individual script. A Fisher's z-transformation was performed to evaluate the statistical significance of the difference between the correlations between the confederate and robot conditions in Script 1, but it was not found to be significant.

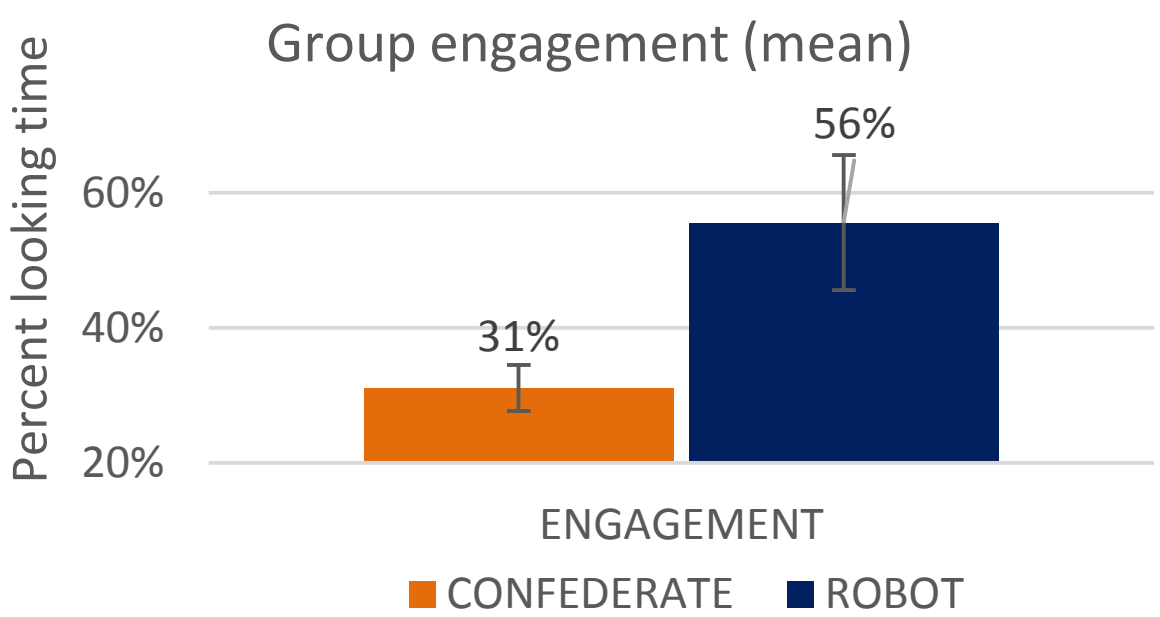

**Figure 6**: Overall engagement was significantly greater in the robot condition compared to the confederate condition.

Nonverbal IQ was significantly negatively correlated with robot engagement ($r=-0.697$, $p<0.05$) but no significant correlation resulted for the confederate condition. A Fisher's z-transformation indicated that the difference between the robot and confederate correlations is significant ($Z = -1.97$, $p<0.05$), and suggests that there may be a connection between lower nonverbal IQ and higher robot engagement.

Two potential explanations for this significant difference are that higher IQ participants needed less time to collect the social cues to respond to each social task or, simply directed their attention more often to other targets in the study space. Lower IQ participants may have spent more time attending to the robot's social cues to improve their task performance or, may have preferred to engage with the robot above other potential targets in the space. In either scenario, while there is a clear connection between IQ and performance in both conditions (Fig. 7), there does not appear to be a strong connection between IQ and engagement (Fig. 8). However, further analyses indicate that greater engagement contributed to higher performance, especially in the robot condition, irrespective of IQ or the order in which the study conditions were presented (Fig. 9). We more fully examine engagement within each study condition and between conditions in sub-section 5.2.

## 5.2 Engagement results

To evaluate frequencies of engagement with social targets and their potential contribution to participant performance, we computed percentages of looking times for each of the four social engagement targets and a cumulative total looking time for all targets for each condition, and for each participant. We first present results from analyses conducted to examine between-condition differences in engagement to ascertain if there is a connection between study condition, IQ and engagement. Next, we discuss target-specific engagement results for each condition.

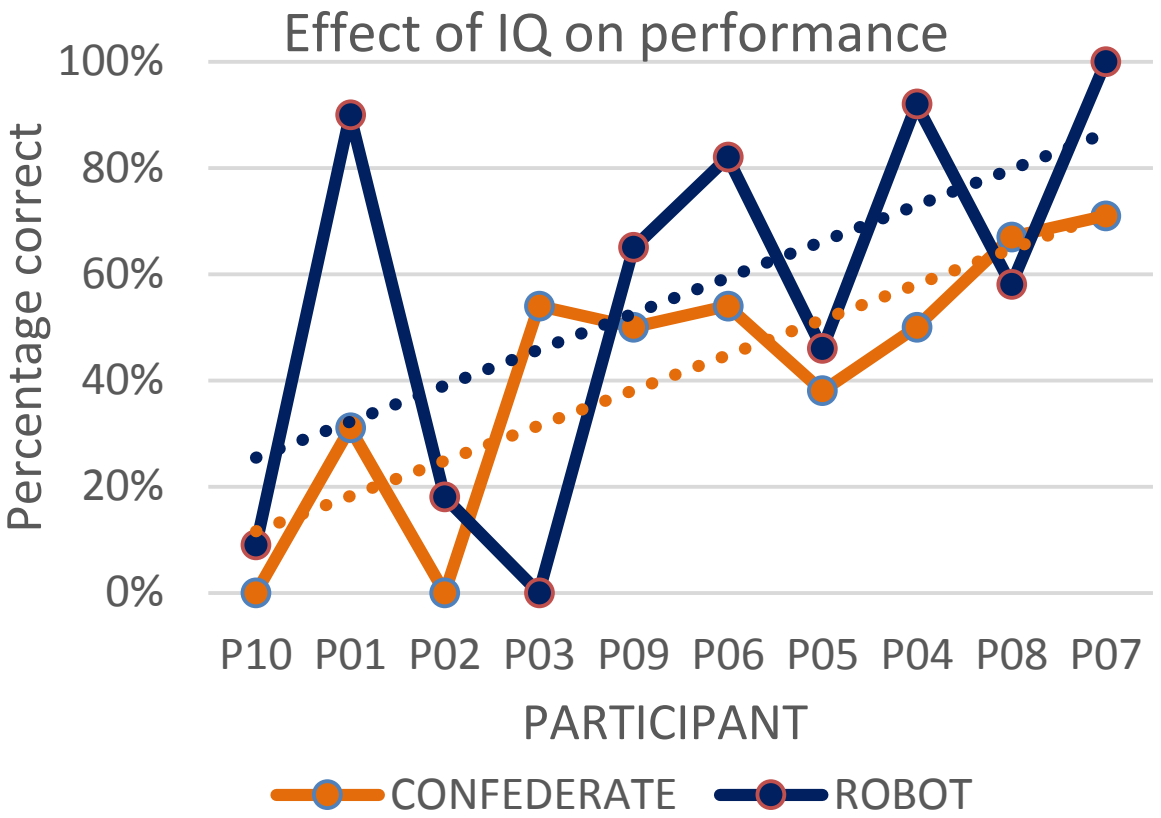

**Figure 7**: Participant IQ is related to performance in both study conditions (IQ ordered from low to high, left to right)

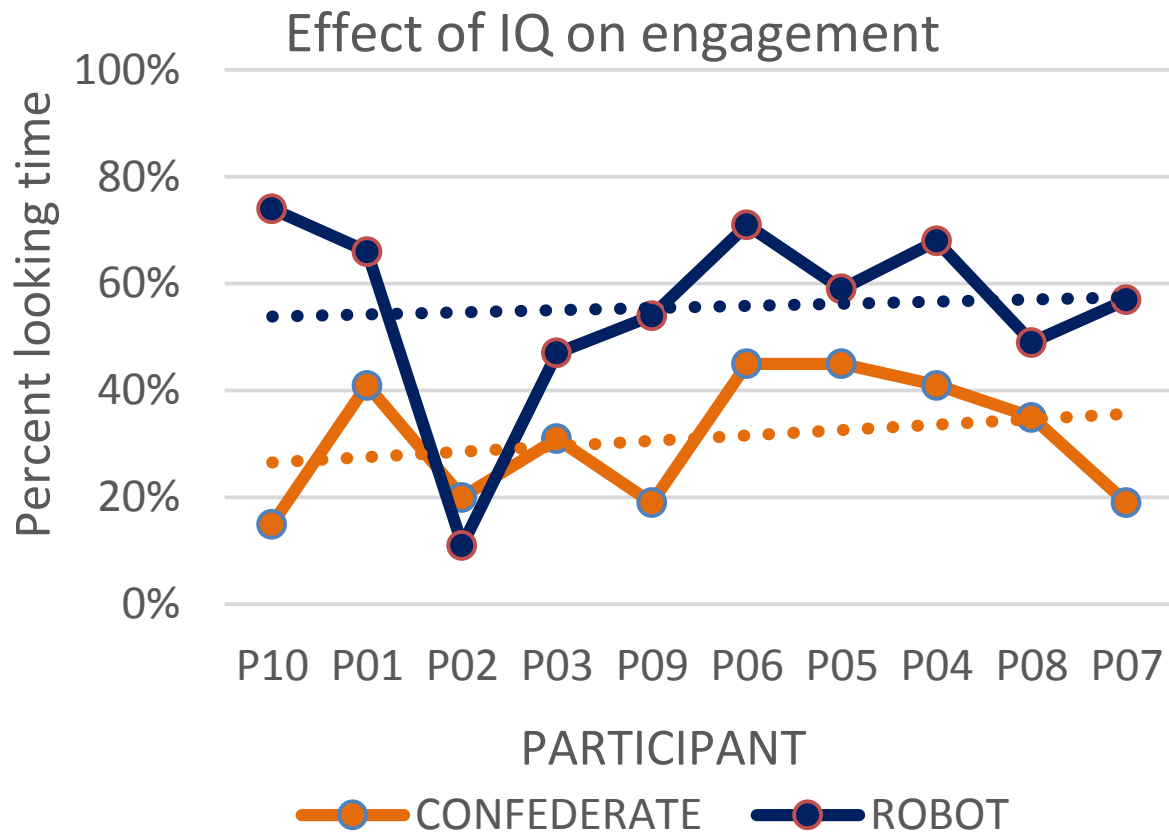

**Figure 8**: Participant IQ is not significantly related to engagement in either condition (IQ ordered from low to high, left to right)

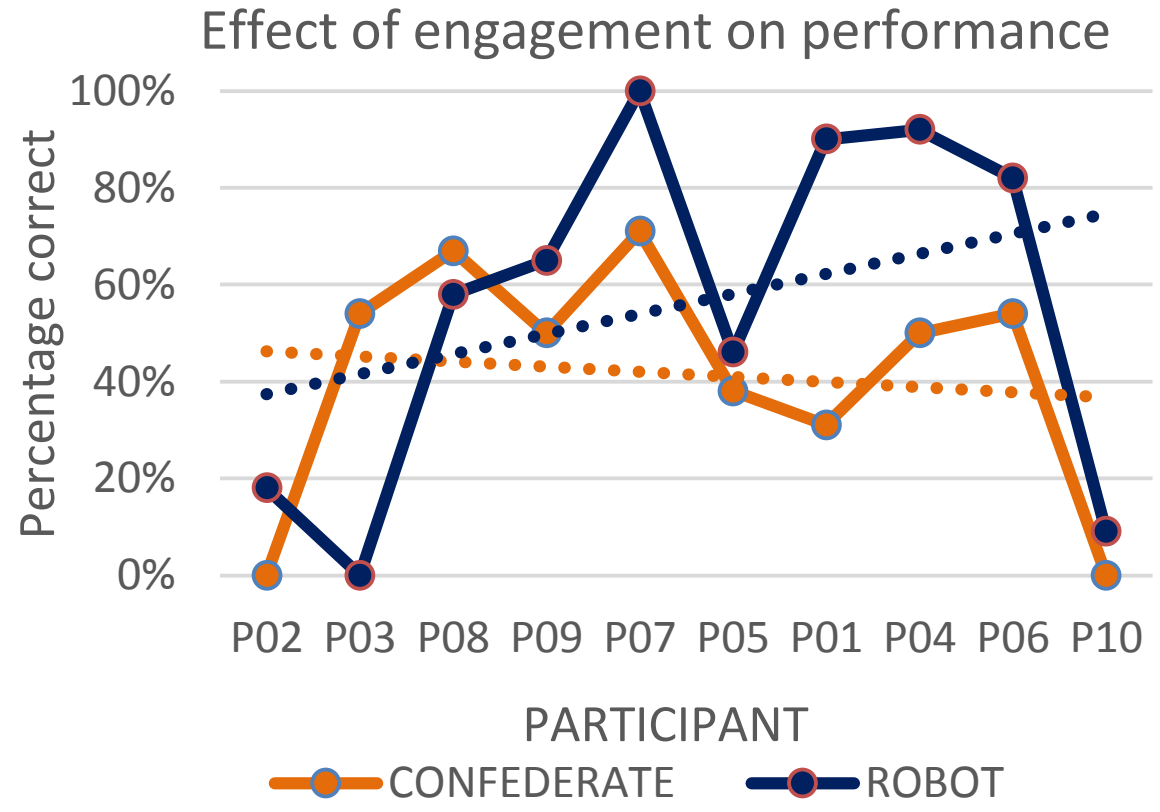

**Figure 9**: Participant engagement significantly contributes to performance in robot condition but not confederate condition (Engagement levels ordered from low to high, left to right)

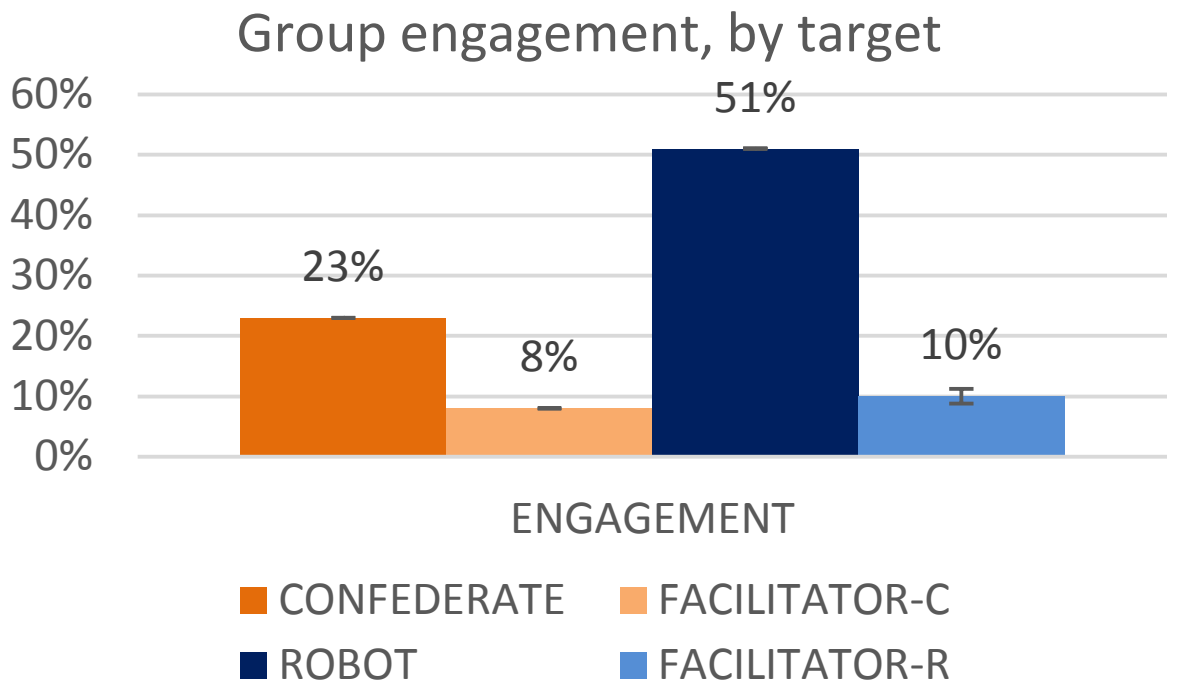

**Figure 10**: Comparing engagement for individual targets. Left: Confederate; Left-middle: Facilitator in confederate condition; Right-middle: Robot; Right: Facilitator in robot condition.

Finally, we evaluate the connection between engagement and performance within the context of each condition.

*5.2.1 Engagement by condition, IQ.*

In the robot condition, overall engagement was approximately 25% greater than in the confederate condition (Fig. 6). Additionally, individual engagement levels revealed that 90% of study participants were overall more engaged with social targets in the robot condition than in the confederate condition. Of the 90% that were more engaged in the robot condition, the average engagement improvement was approximately 30%. For most measures, engagement was not significantly correlated to IQ, except in the robot condition where Nonverbal IQ was negatively associated with robot engagement ($r=-0.697$, $p<0.05$).

*5.2.2 Engagement by target.*

Total confederate engagement was strongly correlated to overall engagement within the confederate condition ($r=0.859$, $p<0.01$) as total robot engagement was significantly correlated to overall engagement within the robot condition ($r=0.692$, $p<0.05$).

Analysis of robot and confederate engagement revealed that 100% participants were more engaged with the robot than the confederate, 60% were more engaged with the facilitator in the robot condition, 20% were equally engaged in both conditions and 20% were more engaged with the facilitator in the confederate condition. Participants spent approximately 23% of their time during the coded phase portions looking at the confederate in the confederate condition. The total duration of time participants spent directing their attention to the confederate varied widely, ranging from 9%-39% of the total coded time. Conversely, the percentage of looking time at the robot was more than twice that duration, averaging approximately 51% for both coded session phases of the robot session. In the robot condition, directed attention varied considerably from 34% to 68% of the total time.

Although engagement with the confederate and robot varied greatly, the percentage of time spent looking at the facilitator in either condition was similar and comparatively low. The mean percentage of time participants spent attending to the facilitator during the confederate condition was 8% while the total looking time at the facilitator during the robot condition was 10%. Within the confederate condition, engagement varied between 2% and 16%, while facilitator engagement in the robot condition ranged from 2% to 31% of total coded engagement time.

Considering that the confederate and the robot were actively engaging each participant within the free conversation, story-telling and scripted portions of each session, it was expected that participants would direct more of their attention to these social targets. However, the significantly greater engagement with the robot compared to the confederate, especially when evaluated alongside the significantly higher task performance in the robot condition, underscores the importance of evaluating the connection between engagement and performance. Because one of the main thrusts of this study was to explore the relative benefits of employing a robot in the role of peer student in this new robot-assisted, learning-by-teaching ASD intervention, we next present results from analyses conducted to examine this potential relationship.

*5.2.3 Engagement and performance.*

Multiple linear regression was conducted to predict overall performance in the robot and confederate conditions from IQ, age and engagement. Neither IQ nor age were singularly, statistically predictive of performance in the confederate or robot condition. Nor were IQ and age jointly predictive of performance in either condition. However, although IQ, age and overall engagement did not collectively predict overall performance in the confederate condition, these variables statistically significantly predicted overall performance in the robot condition, $F(3, 6)=5.399$, $p<0.05$, $R2=0.730$. These findings further emphasize the important and significant connection between engagement and performance, especially in the robot-assisted condition of this study.

These results confirm our expectations that IQ and age contributed to improved performance within this robot-assisted intervention. However, one aspect of these results was not anticipated and may serve to inform our future work with robot-assisted pedagogies. While greater engagement significantly contributed to predicting performance in the robot condition, it did not contribute significantly to performance in the confederate condition.

# 6 STUDY LIMITATIONS

As a preliminary work examining the potential of a novel robot-assisted intervention, we recognize some existing limitations of this study. First, results are based on a relatively limited sample size, therefore caution should be used before generalizing to the larger ASD population. To estimate effect sizes for engagement and performance, we computed a Cohen's *d* for each measure. A Cohen's effect size value for overall engagement ($d=1.6$) suggests high practical significance while the effect size for overall performance ($d=0.48$) suggests medium practical significance.

It is also possible that children were less comfortable correcting an adult than the robot. For all human confederate-assisted sessions, the same adult played the role of the social conversational partner, in order to ensure consistency of delivery. Although having an adult play this role provided a more controlled setting for this study, it is possible that employing an adult in this role may have affected performance. However, given the difficulty of recruiting age-matched peers for this purpose, this potential limitation may actually lend further support to using a socially imperfect robot for this particular intervention.

Finally, in a few instances the video coding of attention was not possible due to a temporary or complete obstruction in the field of view. In these cases, the portions of video where the child's attention to the robot, confederate and/or facilitator were not coded and the total session time was adjusted accordingly.

## 7 DISCUSSION

In this work, we introduce a novel, teaching pedagogy for children with autism. The approach positions the robot in the role of peer student and engages the child as a peer tutor to explore the impact of a robot-assisted, learning-by-teaching pedagogy on performance of a simple social skills task.

*This study is the first research, to our knowledge, to investigate the potential benefits of a robot-enabled version of this classroom pedagogy for students with autism.* There are a number of compelling reasons to engage students with autism in peer tutoring *and* to employ a social robot to fulfill the role of peer student.

**A novel, robot-assisted peer tutoring intervention**. An increasing number of children with autism are attending mainstream schools. The majority of these children receive regular interventions that provide opportunities to practice social skills within group settings, one-on-one intervention with an educator or targeted therapy with a specialist. However, it is still uncommon for a child with autism to have the opportunity to further develop the social skills they have acquired during interventions and foster a deep, generative understanding of these skills by engaging others in peer tutoring. The learning-by-teaching pedagogy offers these children a number of potential social benefits that may facilitate long-term learning and improved outcomes. A growing body of work evidences the long-term benefits of academic peer tutoring for children with autism and this is especially true for lower-achieving students. Although social skills are typically learned and reinforced through interventions like video modelling and group interactions with other peers, the peer tutoring approach for developing social skills in children with autism has not been widely applied.

Despite the numerous benefits associated with the approach, a number of obstacles make the deployment of social skills peer tutoring difficult in practice. Recruiting the appropriate peers to engage in peer tutoring is often difficult, especially when the peer tutor and the peer student should be children from a similar age group and with similar social skills difficulties. Further, providing consistent opportunities to practice the targeted social skills and time to develop the ability to provide appropriate feedback requires practice, making it difficult to enact with two children who may both be learning the skill but possess different levels of proficiency. Employing social robots in the role of the peer student may help surmount these challenges.

**Social robots as peer students**. The complications inherent in engaging peers as students in a social skills learning-by-teaching pedagogy may be remedied by leveraging the unique features of social robots. First, because the social behavior of robots is customizable, the social proficiency of the robot can be adjusted to accommodate the targeted level of interaction for each child. Second, social robots can repeat similar social scenarios without fatigue or frustration and without the negative interpersonal ramifications of peer social interactions gone awry.

This preliminary study employed a social robot in the role of peer student and engaged children with ASD as peer tutors to explore the feasibility and effectiveness of this novel approach for promoting social skills. To this end, we implemented a simple social skills activity wherein participants were tasked with identifying and correcting a limited set of 4 social skills errors. To further investigate the added value of employing a social robot in this role compared to a human social partner, we developed a randomized control trial and randomly assigned participants to one of two study groups: robot-first or confederate-first. For each of the two conditions, engagement and performance measures were recorded and empirically evaluated.

**Significance of results**. Results from this study confirm the feasibility of this new robot-assisted intervention and support the significant potential of this approach for promoting social skills. Performance scores in the robot condition exceeded performance scores in the confederate condition for 80% of participants. Engagement was also significantly improved, with 90% of all participants showing a significantly higher level of overall engagement in the robot condition. Further, although engagement did not contribute significantly to predicting performance in the confederate condition, engagement did contribute significantly to predicting performance in the robot condition. These findings further underscore the attentional value of employing a robot in an autism intervention and the potential therapeutic benefit of employing a social robot as a peer student.

Anecdotally, this approach may also elicit other positive socioemotional behaviors. For example, after making several mistakes and receiving correction, the robot would say, "I'll try to do better next time." This prompted several children to reassure the robot by responding, "It's OK L-E!", "You're doing a good job!" and "It's alright!" These responses were not observed during the confederate condition. The collective impact of these results emphasizes the value of employing robots as peer students in this context while demanding further exploration of the potential of this approach for promoting long-term, social skills learning.

**Future work**. This study engaged a small group of participants with ASD in two peer-tutored scenarios to examine performance and engagement during a simple social skills task. Future research with a larger population of children with ASD will be required to empirically evaluate the utility of the approach for promoting social skills acquisition and long-term learning.

## ACKNOWLEDGMENTS